# How the Advent of Ubiquitous Large Language Models both Stymie and Turbocharge Dynamic Adversarial Question Generation

**Yoo Yeon Sung**
University of Maryland
yysung53@umd.edu

**Ishani Mondal**
University of Maryland
imondal@umd.edu

**Jordan Boyd-Graber**
University of Maryland
jbg@umiacs.umd.edu

## Abstract

Dynamic adversarial question generation, where humans write examples to stump a model, aims to create examples that are realistic and informative. However, the advent of large language models (LLMs) has been a double-edged sword for human authors: more people are interested in seeing and pushing the limits of these models, but because the models are so much stronger an opponent, they are harder to defeat. To understand how these models impact adversarial question writing process, we enrich the writing guidance with LLMs and retrieval models for the authors to reason why their questions are not adversarial. While authors could create interesting, challenging adversarial questions, they sometimes resort to tricks that result in *poor* questions that are ambiguous, subjective, or confusing not just to a computer but also to humans. To address these issues, we propose new metrics and incentives for eliciting good, challenging questions and present a new dataset of adversarially authored questions.

## 1 Introduction

One of the major weaknesses of current QA models come from training crowdsourced datasets including artifacts (Weissenborn et al., 2017; Ettinger et al., 2017; Jia and Liang, 2017). Natural questions (Kwiatkowski et al., 2019) are likewise not immune: they contain idiosyncracies, ambiguities, or false presuppositions (Min et al., 2020, 2022). Thus, recent attempts have focused on dynamically creating adversarial examples to challenge state-of-the-art models (Nie et al., 2020). Kiela et al. (2021) (DADC[1]) invites direct human interaction with the models when human authors write questions to stump models (Wallace et al., 2022; Bartolo et al., 2020).

However, advent of LLMs makes it difficult for humans to stump QA models not only because the models became knowledgeable but also because the models are more opaque (Raffel et al., 2020; Lewis et al., 2020; Bowman, 2023). These disadvantages of LLMs leave the human authors discouraged and untaught on the groundings of why their questions are not successful in fooling the machines.

[1] https://dynabench.org/tasks/qa

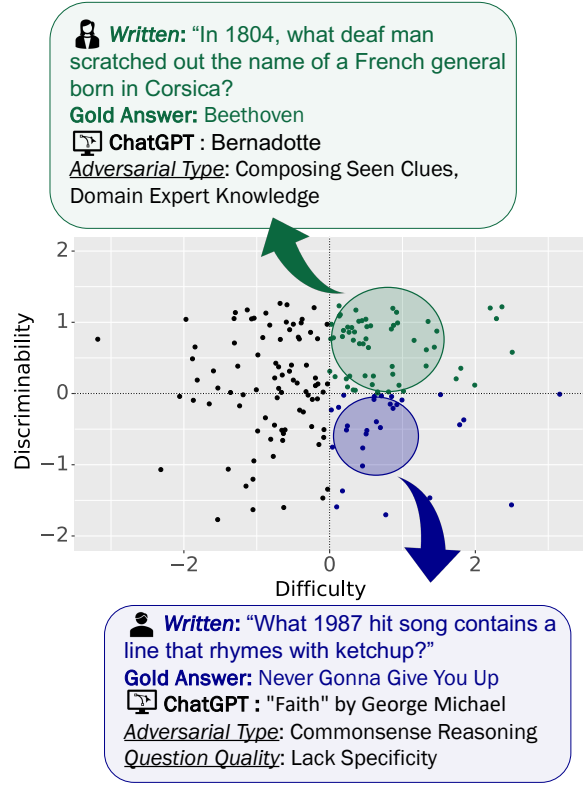

Figure 1: Our IRT analysis exposes what makes for good and poor adversarial questions. The *poor* questions that had *low* discriminability but *high* difficulty (blue) lack specificity despite their adversarialness against CHATGPT (e.g., There are many songs in 1987 containing a line that rhymes with "ketchup"). The questions that had both high discriminability and difficulty (green) met the criterion of being good, adversarial questions.

This paper makes three contributions: an interface *for* DADC *for* LLMs *for* QA (§2 and §4), the first metric for what makes for a good adversarial question (§3), and a characterization of the differential strengths of human vs. computer question answerers circa 2023 (§5).

We collect adversarial questions via an interface designed to examine how LLMS and retrieval models change the dynamics in the DADC. Then, for a clear quality assessment of the collected questions, Section 3 defines a clear, numerical metric of what a *good* adversarial example is based on classic psychometric approaches. The metric requires the answer data from humans and machines; we run a *human VS computer* competition (§3.3). Because we want to account for the range of abilities in populations of the humans and computers that answer questions, we adopt Item Response Theory (Sedoc and Ungar, 2020; Lalor et al., 2019, 2016, IRT) that mathematically models answerers' response data. This metric incentivizes the authors with the best accuracy and encourages them to write better questions. While previous approaches have failed to define what it means to succeed in creating quality adversarial questions, our metric contributes to quantitative adversarial question evaluation.

## 2 Retracing DADC for LLMS

To investigate if and how LLMs impact the adversarial writing process in DADC setting, we present an interactive adversarial question generation interface *updated* with retrieval models and powerful LLMS (Figure 2). Unlike the traditional DADC interface, we include a component that simultaneously displays predictions of LLM-powered QA models. Moreover, to combat the opaqueness of LLMs, we demonstrate different types of retrieval models that yield evidence where the model attained hints for prediction; this way, writers can reason why their questions failed to stump the model and tweak initially written questions (§4). Through our interactions with users, we discovered that LLMs also encouraged a new failure mode of adversarial questions: overly vague questions that neither humans nor computers could answer (Figure 1). These vague questions aligned with errors in current QA datasets: ambiguity (Min et al., 2020), contain false presuppositions (Min et al., 2022) or questionable assumptions (Kim et al., 2022). We had a significantly higher proportion of usable questions when the authors were trained to write good questions for humans.

## 3 What are Good Questions?

Before discussing how to write adversarial questions, we need to define what makes a *good* adversarial question. A naïve explanation of an adversarial example is a question that a human can answer but a computer cannot (Ribeiro et al., 2018). However, this definition is unsatisfying because humans and computers come from *populations*. A more concrete definition should be able to account for the range of abilities in populations of the humans and computers that answer questions (Lord et al., 1968; Hopkins and May, 2013). For humans, a mathematical framework for measuring what makes for effective adversarial questions exists: item response theory (IRT) (Reckase, 1998; Lalor et al., 2016).

### 3.1 IRT models for QA

For QA models, IRT defines the difficulty of a question $j \in \mathcal{J}$ as $\theta_j$ and the skill of a subject $i \in \mathcal{I}$ as $\beta_i$ (Lalor et al., 2019). The higher the subject's skill is compared to the question's difficulty, the more likely the subject is to answer the question. The difference between difficulty and skill is multiplied by a final parameter—discriminability $\gamma_j$—which encodes how effectively the question rewards skill.[2] Thus, good questions have higher discriminability. Taken together, this induces a probability $P_{ij}(r_{ij})$ that subject $i$ will answer the question $j$ correctly, given that $r_{ij}$ is a binary response of a subject $i$ successfully answering question $j$ (Martínez-Plumed et al., 2019):

$$P_{ij}(r_{ij} = 1 \mid \theta_j, \beta_i) = \frac{1}{1 + e^{-\gamma_j(\beta_i - \theta_j)}}. \quad (1)$$

To discover the IRT parameters that best explain the whole data, $\beta_j \in [-1, 1]$, $\theta_j \in [-1, 1]$, and $\gamma_j \in [0, 1]$, we turn to variational inference (Jaakkola and Jordan, 1997) for the full generative process, an effective approximation method for intractable posterior distribution (Natesan et al., 2016).

### 3.2 IRT parameters for good questions

After optimization, we use the learned parameters to recognize the best questions and incentivize the authors. First, the good questions have the largest margin between human and computer difficulty. Given an author $a$, an effective adversarial set of

---
[2]Perfect discriminability means that any subject with positive difference between skill and difficulty will answer the question correctly.

questions $Q_a$ should have a large margin $\mu_a$ between human ($h$) and computer ($c$) difficulties

$$\mu_a = \frac{1}{|Q_a|} \sum_{j \in \mathcal{J}^{(a)}} |(\theta_j^{(h)} - \theta_j^{(c)})|. \quad (2)$$

Second, the best question set should include questions with the highest aggregate discriminability $\kappa_a$, meaning that they distinguish the high-skilled answers and low-skilled answers

$$\kappa_a = \frac{1}{|Q_a|} \sum_{j \in \mathcal{J}^{(a)}} \gamma_j. \quad (3)$$

Third, we want to have variety in the *human* difficulty: some questions should be easier, some should be harder (we avoid questions that every human can answer nor questions that only an expert can answer). Thus, we encourage question sets' difficulty to have as large a median absolute deviation $\delta$ as possible:[3]

$$\delta_a = \text{median}\left(\left|\theta_j^{(h)} - \text{median}_{j \in Q_a} \theta_j^{(h)}\right|\right) \quad (4)$$

We standardize all variables ($\mu$, $\kappa$, and $\delta$) to have zero mean and unit variance. This normalization allows calculating across the variables when creating an incentive metric (§3.3):

$$\mu_a = \frac{\mu_a - avg_\mu}{std_\mu}. \quad (5)$$

### 3.3 Incentive metric for question evaluation

To make use of IRT for evaluation, we hold a *Human VS Computer Competition* with the written questions. We follow a competition format[4] where participants write questions and answer others' questions themselves (Jennings, 2007). This rewards answering questions and incentivizes writing good questions, as your peers hear and judge your questions. For the answering competition, we also invite the non-question writers, computer system submissions and LLM-based QA systems (e.g., T5 and Distilbert[5]). After each cycle, we reward the answerer team whose accuracy of the answers was the highest (with the highest skill $\beta$) by:

$$a_{\beta^*} = \arg\max_a \beta_a \quad (6)$$

Then, to reward the writer team, we score each question set by summing their difficulty margin,

---

[3] The computer difficulty should be as high as possible.
[4] https://acf-quizbowl.com/packet-submission-guidelines/
[5] Both finetuned on SQUAD (Rajpurkar et al., 2018)

discriminability, and divergence scores. We anticipate these scores to serve as an incentive mechanism for the participants as well as question evaluation.

$$\text{Score}_{Q_a} = |Q_a| \frac{(\mu_a + \kappa_a + \delta_a)}{3} \quad (7)$$

### 3.4 Aiming for adversarial, yet good questions

Our goal in crafting adversarial questions lies in probing the ability of the QA models (e.g., is it robust to adversarial attacks?) rather than satisfying the users who use QA models with information-seeking purposes (Rogers et al., 2023). In Section 4, we propose an interface to help authors to write these questions. However, there is another interface that is unescapable in 2023: CHATGPT. Authors will also use it to help write questions; these questions might ostensibly satisfy the above metrics but are vague or confusing (Figure 1). Thus, we introduce a second constraint: they must satisfy the "norms" of trivia questions, which Boyd-Graber and Börschinger (2020) argues also creates good QA training data.

To this end, we further demand that questions pass vetting by filtering those that lack specificity and factuality, and avoid having many answer spaces and subjectivity (More details in appendix 7). Also, we urge simple-formatted questions distinct from pyramidal questions that have complex forms and are likely to be less used in the real world (Boyd-Graber and Börschinger, 2020; Wallace et al., 2019). For example, a good question is *"What is the post-apocalyptic science fiction action film directed by a Korean director but not by the director of Parasite is about the class struggles of passengers on a train attempting to survive their journey?"*; it stumps the model with a multistep-reasoning tactic, while being specific. Acknowledging that powerful LLMs (e.g., CHATGPT) are omnipresent in applications today, we build an interface that implements similar models to study their influence in adversarial writing.

## 4 Question Writing Interface

Our interface (Figure 2) focuses on using external content (e.g., other than given passage) when the retrieval models obtain evidence for their prediction. This allows users to access diverse and resourceful information when writing questions that do not overlap with the evidence. Moreover, we incorporate LLM guidance to stress-test the impact of LLMs in dynamic natural-like adversarial question generation. Encountering our interface main page, the authors first select the Wikipedia Page and enter the *question writing* browser. The title of the Wikipedia page is the *target answer* they will write their question on. We introduce our *real-time* machine-stumping mechanisms in our interface.

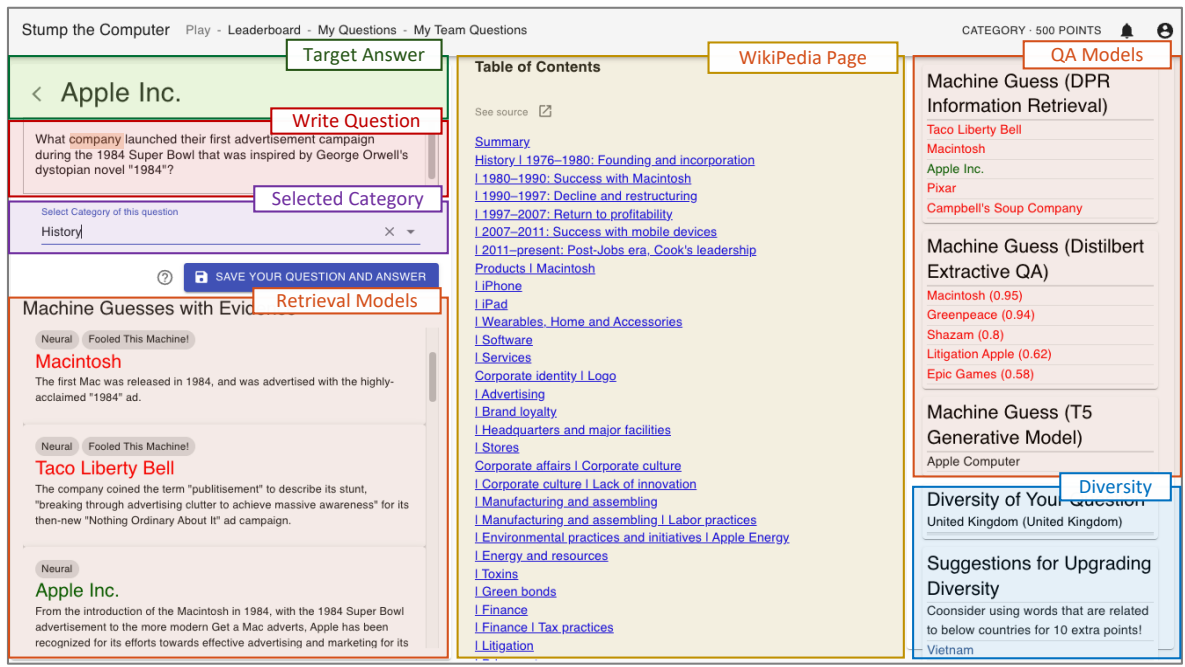

Figure 2: As the target answer to the question shoud be "Apple Inc", the interface is updated with answers from retrieval models with the most relevant sentence and from LLMs (e.g., Distilbert, T5). Also, the highlights are updated by the input perturbation technique. The diversity widget is updated with the country representation of the questions and suggested countries.

### 4.1 Stumping the models

As an author reads the answer and the context (e.g., Wikipedia title and content) and writes questions, the retrieval widget (Figure 2 bottom left) and QA models widgets (right) are updated (Eisenschlos et al., 2021). Motivated by Feng et al. (2018), we embed the input perturbation inside the question writing widget (top left) to highlight which words trigger model predictions; (*company* is considered as the most important token to change the prediction).

**Retrieval Systems** We use sparse and neural retrieval models: the TF-IDF (baseline) model and Dense Passage Retrieval (Karpukhin et al., 2020, DPR). To ensure the contemporaries of the retrieval systems, we created the databases of retrieval systems from the latest Wikipedia pages to follow up-to-date information (Appendix 10.11). We then use the RoBERTa-based FARMReader—Finetuned on SQUAD—to read and sort the retrieved sentences and titles by their relevance. These are the passages that the systems are using to answer; authors can rephrase the question to avoid retrieving the information or to prevent the reader from extracting the correct answer. In Section 5.3, we investigate whether this generalizes to more complicated models like LLMs. Also, we instruct the authors to revise their questions if the target answer appears at the top of the retrieval widget, meaning they failed to stump the model. If the answers do not match the target answer, the answer is tagged with `Fooled This Machine.`

**LLM-based QA Systems** We enrich the model guidance by using both extractive and generative model answer predictions. For extractive QA, we use fine-tuned DistilBERT[6] (Rajpurkar et al., 2016) and use the same Wikipedia database as DPR. Since we value the interaction between the authors and the models, we take advantage of its promptness and lightness. Moreover, we use Google's T5 to answer the human-authored questions in a closed-book setting (Raffel et al., 2020).[7]

---
[6] fine-tuned on SQUAD
[7] We do not include CHATGPT as a stumping technique because of its latency, but as discussed in §3.4, we know authors use it.

### 4.2 Topic Diversity

Apart from stumping the models, we encourage topic diversity in the questions (Wang et al., 2020). We ask the authors to submit their question packets with a fixed number of each category from Art, Literature, Geography, History, Science, TV and Film, Music, Lifestyle, and Sport (Appendix 10.4).

### 4.3 Interface Incentive

To encourage competition and authors to monitor their progress, authors can monitor how many questions they wrote per category and their diversity level on the `Writer Leaderboard` (Appendix 10.13). Once the authors finish writing the questions, the `Machine Leaderboard` updates whether their questions stumped CHATGPT.

## 5 Results: Are they Good Questions?

This section evaluates and analyzes the questions written in the interface designed to target LLM-powered QA models (e.g., T5 and DistilBert); we explore 1) the question quality with our incentive metric 2) what kind of adversarial techniques appear in *good* and *poor* questions 3) how questions that stump the LLMs in the interface generalize when applied to more complex LLMs (e.g., CHATGPT), and 4) how explanations from retrieval models help in constructing an adversarial proxy to stump the LLMs. To make this feasible, we hold two rounds of competition with 12 author teams and 12 answerer (human and machine) teams. In answerer teams, there were eight human teams[8] and four machine answerers (e.g., DPR, T5, DIS-TILBERT, and CHATGPT). We collected 399 adversarial questions through the interface and ran a competition with 184 edited questions.

We noticed that there was an improvement in question quality when authored by Trivia writers acquainted with "trivia norms" (Boyd-Graber and Börschinger, 2020; Rodriguez and Boyd-Graber, 2021). The question sets written by Trivia experts contained 57% acceptable questions, while those written by college students contained 38%. One failure mode of vague questions was "What video game movie featuring one of the *most popular* and *well-known* icons in video games stars Chris Pratt and Jack Black?"; the question was too subjective.

---

[8]The authors and human teams were formed by volunteers from the Trivia Community and college students.

### 5.1 Questions written against LLMS

**Difficulty** To check the adversarial-ness of our questions, we assess how the collected questions have a high margin between human difficulty and machine difficulty. Table 1 demonstrates the number of questions that stumped the human and machine teams. The number of questions that stumped *some* humans and *all* machines were the highest. Also, the ratio of correctly answering the questions from the most competent human team to the computer team was 79 to 21, suggesting that most of the questions were adversarial.

| | | Number of Questions Stumped | |
|---|---|---|---|
| | | MACHINE | |
| | | All | Some |
| HUMAN | All | 73 | 8 |
| | Some | **90** | 13 |

Table 1: The number of questions that stumped *some* humans and *all* machines were the highest: 90 adversarial questions.

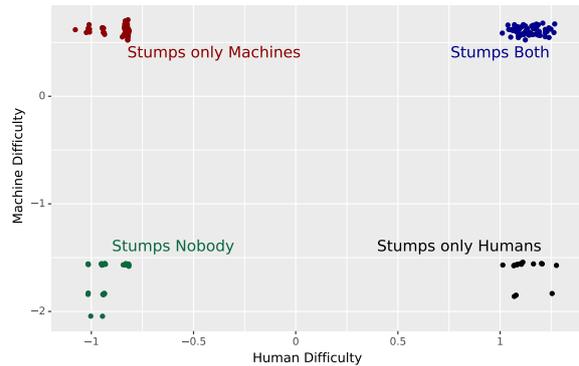

Figure 3: The number of questions that stump only machines (top left) was comparable with the number of questions that stump both humans and machines (top right).

Figure 3 shows four clusters of questions according to the $\theta_j^{(h)}$ and $\theta_j^{(c)}$ learned from the IRT models. The red cluster (top left) denotes the questions that stump only the machines (39%), the blue cluster (top right) denotes the questions that stump both humans and machines (36%), the black cluster (bottom right) denotes the questions that stump only the humans (13%), and green cluster (bottom left) denotes the questions that are easy (11%). We infer that our framework helps create adversarial questions.[9] Table 2 illus-trates the examples that have the highest margin between human difficulty and machine difficulty per cluster.

---

[9]The questions that stumped both do not necessarily mean

| Stumped Subjects | Question | Answer |
|---|---|---|
| Only Machines | A German admiral sailing for Russia named what islands for an English captain and not for the librettist of the HMS Pinafore nor for the announcer of Jeopardy! | Gilbert Islands |
| Machines and Humans | What color did Real Madrid wear during the 2017 Champions League final? | Purple |
| Only Humans | Which of the first Adidas Yeezy Boost 350 designs had an out of this world themed name? | Moonrock |
| Easy | What famous art piece that is currently in France is referred to as La Giaconda? | Mona Lisa |

Table 2: From each cluster, we display examples that had the highest margin between the human and machine difficulty.

**Discriminability**  Also, we check the discriminability value ($\gamma$) of the collected questions. Table 3 shows the questions sorted by $\gamma$. The questions with the highest values were only answered by teams familiar with literature and history; the adversarial tactic of *Domain & Commonsense knowledge* was used so that the question could be only answered by models trained on specific knowledge. Moreover, high discriminability data points are scattered across all difficulty levels, meaning that questions at all difficulty levels mostly rewarded the answerer's skill (Figure 1).

### 5.2 What kind of adversarial tactics do LLMs invoke and what makes these less *good*?

For a deeper analysis, we scrutinize what kind of adversarial tactics were used by writers to stump LLMs and evaluate if they are good or poor. To understand *how* they are poor, we examine if there exists any correlation between discriminability and the adversarial tactics the question used.

We manually tag the questions with problematic question types and adversarial types. We added more adversarial types to Wallace et al. (2019),

---

that they are not adversarial. These questions may not be as easy as the writers intended them when writing, as writers and answerers may have different knowledge or answering skills.

| Question | Answer | $\gamma$ |
|---|---|---|
| What city in England is home to the studios where the British game show that resembles quiz bowl is filmed and where the test of whether an AI is intelligent was proposed? | Manchester | 0.193 |
| The 1973 Thomas Rockwell novel for children, about a school boy who loses a gross dare, was written in America. However, a similarly-titled religious edict published in 1521 hails from this old European city of about 80,000. | Worms, Germany | 0.192 |
| ⋮ | ⋮ | ⋮ |
| Who was able to turn men into stone sculptures by just taking a glance at them? | Medusa | 0.003 |
| Which political party governs the country directly south of Botswana? | African National Congress | 0.003 |

Table 3: Questions sorted by discriminability($\gamma$) value of the IRT model.

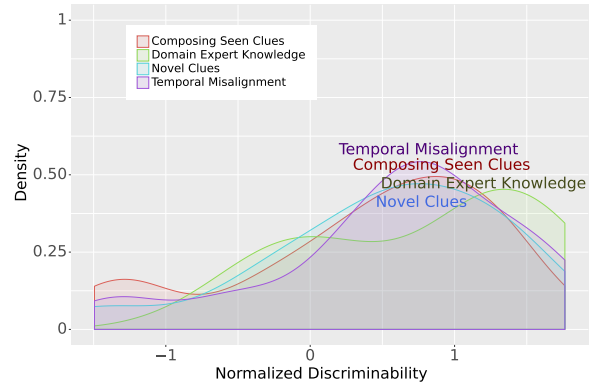

Figure 4: The adversarial techniques *Temporal Misalignment*, *Composing Seen Clues*, *Domain Expert Knowledge*, and *Novel Clues* are used more frequently in questions with high discriminability.

such as *Domain Expert Knowledge*, and added *Subjectivity* to Boyd-Graber and Börschinger (2020) (Appendix 10.5 and 10.6). Consider the question "What is the piece of clothing worn underneath the shirt?"; in some cultures, "undershirt" would be the correct answer, while "pants" is acceptable for others. In this case, the question type is *subjectivity* and *Lacks specificity*. Likewise, a single question can be poor in different ways (examples in Appendix 10.7).

From these annotations, we investigate the cor-relation between what kind of adversarial types appear within questions of different levels of discriminability. We find that adversarial writing techniques, such as *Temporal Misalignment*, *Composing Seen Clues*, *Domain Expert Knowledge*, and *Novel Clues* often appear in questions with high discriminability (Figure 4). On the other hand, *Logic&Calculation* technique is used in questions of all discriminability levels (Appendix 5).

### 5.3 Are questions written against retrieval models generalize to LLMs?

We then analyze the impact of retrieval models on the written questions and examine if these questions stump the LLM, as our pieces of evidence are not created from the LLMs but retrieval models. We use CHATGPT as an upper bound for questions to stump and to compare retrieval models; CHATGPT performs the best among other LLMs (40% accuracy). To better understand if and how retrieval models were utilized during authoring, we first compare the accuracy of retrieval models to CHATGPT. The proportion of questions that stumped both retrieval model and CHATGPT was the largest (65%), suggesting that authors who targeted retrieval models to write their questions also stump the CHATGPT (Table 4). Among the questions that stumped both models, the accuracy of T5 and Distilbert were 4%, and 0%, respectively, which also hints at T5's support in writing stumping questions.

|  |  | DPR | |
|---|---|---|---|
|  |  | Correct | Incorrect |
| CHATGPT | Correct | 2% | 32% |
|  | Incorrect | 1% | **65%** |

Table 4: Percentage of questions stumping both DPR and CHATGPT were the highest among all, indicating that the questions that were written with DPR guidance often stump CHATGPT.

### 5.4 How do retrieval model evidence help stump LLMs?

**Utility of evidence** Following the analysis on how retrieval model-guided questions generalize against LLMs, we examine the efficacy of retrieval model *evidence* when stumping the LLM. Here, we use CHATGPT as a human proxy to contrast *which* retrieval model's evidence helped CHATGPT to answer (Guo et al., 2023; Mondal et al., 2023). We examine the answers based on these evidence to check if it was successful in stumping CHAT-GPT (Ma et al., 2023; Liu et al., 2023). For each question, we retrieve evidence from both the baseline retrieval model and DPR, and prompt CHATGPT.[10] The CHATGPT will return the answer and whether the evidence helped it to answer. We assign (scores) to the evidence in three rubrics: unhelpful evidence + incorrect answer (0), helpful evidence + correct answer (1), and helpful evidence + incorrect answer (2). Score (2) indicates that the question with *the evidence* stumped the CHATGPT model. Then, we averaged the scores (baseline retrieval: 0.22, DPR: 0.32), finding that DPR's evidence is 10% more helpful in writing questions that stumped CHATGPT.

**Error Analyses with Evidence** It is reassuring that retrieval-based approaches seem to generalize to "closed-book" models. We next look at how the evidence generalize from retrieval models to advanced models like ChatGPT and vice versa.

To retrieve CHATGPT explanation for each question, we prompt[11] it to return an answer and its groundings. The comparison examples are demonstrated in Table 5. First, ChatGPT does not provide good explanations of *why* it answers something when it was incorrect, suggesting that spurning such models' explanations in our interface was wise: it hallucinates "Brad Pitt" slapping someone at an awards show. But when provided the relevant context from retrieval models with access to more up-to-date material, it gets the answer correct, a la retrieval-augmented QA systems like Siriwardhana et al. (2023). This suggests that it would also get the correct answer if it were updated. However, ChatGPT gets the answer correct when its explanations are similar to more traditional retrieval systems, partially explaining the outcome overlap we discovered in Section 5.3 (More examples in Appendix 10.10).

## 6 Related Work

Recently, the NLP community has attended to whether the models trained on benchmarks learned to solve tasks in robust and generalizable ways (Ribeiro et al., 2020). Revealing the models' weaknesses in Bartolo et al. (2021); Nie et al. (2018); Gururangan et al. (2018), researchers have been collecting data within dynamic adversarial generation framework where humans create examples against the target model (Ma et al., 2021; Kiela et al., 2021).

---

[10] We prompt CHATGPT, *Do you think that the question can be answered using this evidence from Wikipedia. If yes, predict an answer using the evidence in one or two words.*

[11] We prompt ChatGPT, *"Answer the question in one or two words and provide an explanation for your answer."*

| | | Explanations | | |
|---|---|---|---|---|
| **Question** | **Answer** | **Baseline** | **DPR** | **CHATGPT** |
| What is the name of the American actor who stood up for his wife with a "slap that was heard around the world" during a popular awards show? | Will Smith | *Correct Answer: Will Smith* "Willard Carroll Smith II (born September 25, 1968), also known by his stage name The Fresh Prince, is an American actor, rapper, and producer." | *Correct Answer: Will Smith* At the 2022 Academy Awards ceremony, shortly before winning, Smith faced public backlash for slapping and shouting at Oscar presenter Chris Rock after Rock made a joke referencing Smith's wife, Jada Pinkett Smith. | *Incorrect Answer: Brad Pitt* The incident of Brad Pitt standing up for his wife with a slap during a popular awards show was widely reported in the media and is a well-known event in popular culture. |
| What is the name of a famous surrealist painting which originally was thought to be inspired by a scientific theory about space-time but was in fact inspired by a soft, surface-ripened, brie-like cheese melting in the sun? | The Persistence of Memory | *Correct Answer: The Persistence of Memory* Another insect that is present in the painting is a fly, which sits on the watch that is next to the orange watch. | *Incorrect answer: A Sunday Afternoon on the Island of La Grande Jatte* Inspired by optical effects and perception inherent in the color theories of Michel Eugène Chevreul, Ogden Rood and others, Seurat adapted this scientific research to his painting. | *Correct Answer: The Persistence of Memory* I predicted this answer because The Persistence of Memory is one of the most famous surrealist paintings and is often associated with the concept of time. However, it is also known that the painting was inspired by Salvador Dali's observation of a melting Camembert cheese in the sun. |

Table 5: Example of evidence that TFIDF, DPR, and CHATGPT provide for their respective answers. The first example shows an example where retrieval models are not stumped but LLMs are. The explanations show that DPR and CHATGPT are pointing to the right claims to predict their answers, and it seems like the specificity is lacking in the question that induced the CHATGPT to be stumped. On the other hand, the DPR evidence in the second example was close (has several overlapping spans), but is stumped with the multistep reasoning technique. CHATGPT captures this nuance and correctly answers the question. **When we added each retrieval model's explanation, CHATGPT corrected itself** and answered "Will Smith".

Recently, Tedeschi et al. (2023) postulate that many *superhuman* models may have a false sense of accomplishment due to poor annotated datasets and biases embedded in the evaluation process (e.g., fixed test sets). Fostering our adversarial dataset creation framework could not only help the experts to create the next generation of data, but also systematically probe models to understand their capabilities (Bowman, 2023; Yuan et al., 2023).

Turning to dynamic adversarial generation for QA, Bartolo et al. (2021) uses a synthetic generation method to create human adversaries. Also, Sheng et al. (2021) introduced a benchmark where the humans interact with existing Visual QA model, and for each image, find an adversarial question. Wallace et al. (2019) and Eisenschlos et al. (2021) both uses dynamic incentive mechanism to create adversarial questions.

Moreover, critics complain that the current evaluation treats each subject independently rather than considering relative differences. To remedy this, Lalor et al. (2019) introduces the IRT ranking method. Followingly, Rodriguez et al. (2021) redesigns the leaderboard framework with a Bayesian leaderboard model where latent subject skill and latent item difficulty predict correct responses.

## 7 Conclusion and Future Work

To test the impact of LLMs to dynamic adversarial question generation, we implement an LLM-powered writing interface with retrieval models. While it is critical for the general public to understand the limitations and abilities of LLMs, our dataset helps probe them and communicate with them in accessible ways. Adversarial datasets help evaluate the gaps in the abilities of LLMs to know where to improve the models in the future and know when we have succeeded. For future work, we plan to include perpetually evolving LLMS in the loop in dynamic adversarial creation.

Moreover, we plan to develop a new rewarding system that not only compensates for good adversarial questions but also those that help in good calibration of the trained models. We suggest building QA models with updated training and test sets of adversarial questions and performing comparative IRT analyses for a fair evaluation of questions.

# 8 Limitation

Even though we empirically validate the quality of our questions, we still struggle to address the concern that examples in out-of-distribution datasets, created from adversarial filtering process. This may lead to modeling that simply obscures the errors that can be made from adversary datasets, rather than fundamentally understanding the patterns that models have problems learning (Bowman and Dahl, 2021). In the future, it would be worthwhile to collect questions by measuring and comparing the difficulty of each of the several machines to stump and thereby improve the generalization of the model.

Furthermore, despite our attempt to gather questions that include demographic diversity, we could not observe any significant changes in the country distribution in the written question (Appendix 10.12). We plan to improve the "Diversity" widget by augmenting missing entities that could be linked to a country's representation.

# 9 Ethical Considerations

We address ethical considerations for dataset papers, given that our work proposes a new dataset. We reply to the relevant questions posed in the ACL 2023 Ethics FAQ.[12]

Our study was pre-monitored by an official IRB review board to protect the participants' privacy rights. Moreover, the identity characteristics of the participants were self-identified by the workers by completing the task.

Before completing the task, we display consent forms for the workers to agree that their answers would be used for academic purposes. They were invited to participate in the writing and answering task for entertaining and academic purposes.
We emphasize the scale and the impact of our research in that it provides the resource and an evaluation metric, not constrained to QA, to resolve the current hallucinations and artifacts in NLP datasets.

---
[12]https://www.acm.org/code-of-ethics

## 10 Appendix

### 10.1 Tasker's Goals in Dynamic QA Generation

When tasking human authors with adversarial writing of questions, Wallace et al. (2019) emphasizes the importance of "who" the authors should be: *talented and eager* question writers with *specific goals*; they should aim to generate questions that stump computers but seem normal enough for humans to answer. To make this work, they recruit members of the Quizbowl community who have deep trivia knowledge and craft question for Quizbowl tournaments (Jennings, 2007). However, their challenge was to convey what is "normal" to authors and stimulate examples that can elucidate the weaknesses of QA models.

### 10.2 Merging Trivia Question Generation and Dynamic Adversarial Generation Process

Many QA datasets are now too easy for modern models as models have become more powerful (Rogers et al., 2023). However, even these easy QA datasets have serious data flaws (Min et al., 2020; Yu et al., 2023), which suggests that creating question-answer pairs is a very challenging task. This is also a norm for questions written for human players, where more than 100,000 questions are produced annually. To create effective and challenging enough questions, the professional experts (e.g., writing staff) take a rigorous editing pass on the questions to decide whether they are adequate enough to guarantee players a fair game (Lelkes et al., 2021; Pollard, 2006). They follow strict guidelines to be selected to be used in the quiz matches. We propose to merge the above pipelines to help improve data creation for robust QA models by adding an editing step to ensure that grammatical errors and nonfactual questions (following the norms of Trivia questions) do not exist in the pool.

### 10.3 Variational Inference for IRT models

To discover the IRT parameters that best explain the whole data, difficulty $\theta_j$ and discriminability $\gamma_j$, we turn to variational inference for the full generative process, an approximation method for intractable posterior distribution in Bayesian inference (Natesan et al., 2016; Lalor et al., 2019). The parameters $\theta$ and $\beta$ follow Gaussian prior distributions and make inferences through joint posterior distribution $\pi(\theta, \beta|Y)$ (Natesan et al., 2016).

### 10.4 Topic Categories of Questions

We ask the question writers to tag their questions with the categories below. With reference to specific categories and examples, we encourage them to be as creative and diverse as possible when authoring the questions. In the interface, they can monitor how many questions they wrote per category. They are required to submit packets with a specific amount of questions in each category.

### 10.5 Question Type Annotation

In Table 7, we list the problematic question types that we ask the annotators to annotate. These are illustrated with descriptions and examples to help them better understand each question.

### 10.6 Adversarial Type Annotation

In Table 8, we list adversarial types (techniques) to determine how each question is using them to stump the models. The annotators are given the description and examples to better understand the reasons why the models may have been stumped. They are expected to tag the examples with the model prediction and question.

### 10.7 Question Examples Annotated with Question and Adversarial Types

Table 9 shows question examples that are annotated with question and adversarial types. The highlights in the question correspond to either adversarial types or question types that are highlighted with the same color.

### 10.8 Correlation between Adversarial Types and Discriminability

We scrutinize what kind of adversarial tactics were used by writers to stump LLMs and evaluate if they are "good" or "bad". To understand *how* they are bad, we examine if there is any correlation between the adversarial-ness the question has and how *good* they are. Figure 4 and 5 shows that *Temporal Misalignment*, *Composing Seen Clues*, *Domain Expert Knowledge*, and *Novel Clues* are used more frequently in questions with high discriminability. On the other hand, *Multistep Reasoning*, *Domain Expert Knowledge*, and *Logic & Calculation* are used less in questions with high discriminability.

### 10.9 Examples sorted by Difficulty Score

In Table 10 and 3, we demonstrate examples sorted by the learned variables difficulty ($\theta$) from IRT

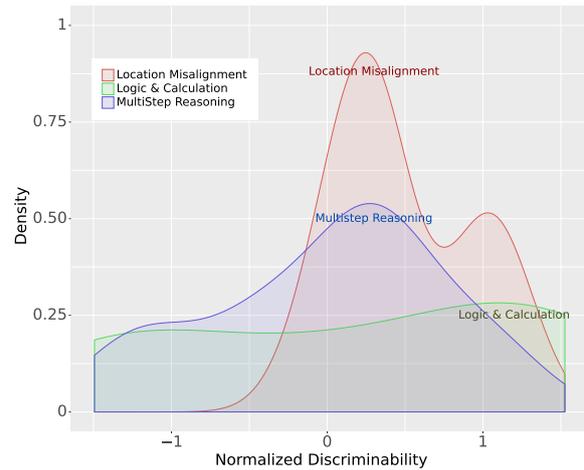

Figure 5: The adversarial techniques *Location Alignment*, *Multistep Reasoning*, *Domain Expert Knowledge*, and *Logic & Calculation* are used less in questions with high discriminability.

model. The examples with the highest variable value is ranked 1.

### 10.10 Explanation Examples from Retrieval Models and CHATGPT

In Table 11, we demonstrate the explanations from retrieval models and CHATGPT models to deeply analyze how explanations from retrieval model may help stump the CHATGPT.

### 10.11 Retrieval System Details

To ensure that the retrieval results help in obtaining up-to-date information for the writers, we created the database for Wikipedia pages and DPR training data. DPR retrieves the most relevant sentence from a database that consists of the Top 1000 popular Wikipedia pages[13] from 2021 to 2022. DPR is finetuned with the 2018 and 2021 QANTA datasets (Rodriguez et al., 2019). For training, we used the questions and gold evidence as positive samples, and sentences from pages that are two hops away (pages linked by randomly selected hyperlinks in the summary section) from the question page as negative samples.

### 10.12 Demographic Diversity Results

We added a "Diversity" widget that determines the entities[14] (e.g., George Orwell) that capture the nationalities[15] (e.g., United Kingdom). We then provide suggestions to the authors to include entities from underrepresented countries. However, the questions' demographic diversity distribution did not conform to the population distribution (Equation 2.2), and the entities in the questions showed few country representations.

---
[13]https://pageviews.wmcloud.org/topviews/?project=en.wikipedia.org&platform=all-access&date=last-month&excludes=
[14]https://cloud.google.com/natural-language/docs/analyzing-entities

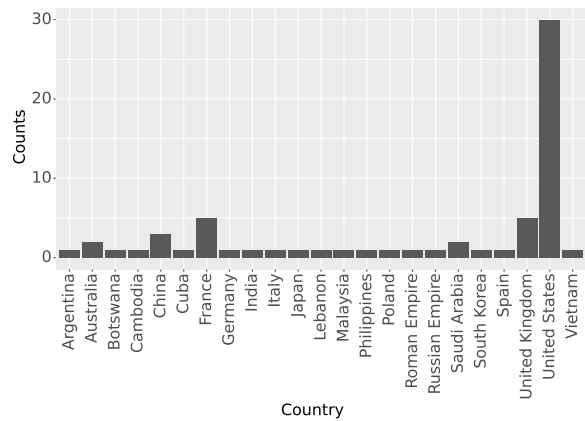

Figure 6: The demographic diversity distribution had negative result as the questions did not contain much nationalities and thus did not conform to population distribution.

## 10.13 Interface Leaderboard

We also build a leaderboard page for writers to keep track of their scores and their diversity score. Figure 7 shows an example of the leaderboard where it displays each writer's name, score, and diversity score.

---

[15]https://www.wikidata.org/wiki/Wikidata:REST_API

| Question | Answer |
| --- | --- |
| Art | Questions about works: Mona Lisa, Raft of the Medussa, B) Questions about forms: color, contour, texture, C) Questions about artists: Picasso, Monet, Leonardo da Vinci, D) Questions about context: Renaissance, post-modernism, expressionism, surrealism |
| Literature Movement | A) Questions about works: novels (1984), plays (The Lion and the Jewel), poems (Rubaiyat), criticism (Poetics), B) Questions about major characters or events in literature: The Death of Anna Karenina, Noboru Wataya, the Marriage of Hippolyta and Theseus |
| Literary Movement | A) Cross-cutting questions (appearances of Overcoats in novels), B) Common link questions (the literary output of a country/region) |
| Geography | A) Questions about location: names of capital, state, river, B) Questions about the place: temperature, wind flow, humidity |
| History | A) When: When did the First World war start?, B) Who: Who is called Napoleon of Iran?, C) Where: Where was the first Summer Olympics held?, D) Which: Which is the oldest civilization in the world? |
| Science | Questions about terminology: The concept of gravity was discovered by which famous physicist?, Questions about the experiment, Questions about theory: The social action theory believes that individuals are influenced by this theory. |
| TV and Film | Quotes: What are the dying words of Charles Foster Kane in Citizen Kane?, Title: What 1927 musical was the first "talkie"?, Plot: In The Matrix, does Neo take the blue pill or the red pill? |
| Music | Singer: What singer has had a Billboard No. 1 hit in each of the last four decades?, Band: Before Bleachers and fun., Jack Antonoff fronted what band?, Title: What was Madonna's first top 10 hit? |
| Lifestyle | Clothes: What clothing company, founded by a tennis player, has an alligator logo?, Decoration: What was the first perfume sold by Coco Chanel? |
| Sports | Known facts: What sport is best known as the 'king of sports'? <br> Nationality: What's the national sport of Canada? <br> Sport player: The classic 1980 movie called Raging Bull is about which real-life boxer? <br> Country: What country has competed the most times in the Summer Olympics yet hasn't won any kind of medal? |

Table 6: Categories of questions along with the subcategories and corresponding examples.

| Question Type | Description | Examples |
| --- | --- | --- |
| Lacks Factuality | Requires information is factual | "Trump, the first woman president of the United States, is charged against federal laws" is non factual as the gender of Trump is male |
| Lacks Specificity (False Presupposition) | Requires more information to be answered with clarity | 'What is the color of Flamingo's feathers?' is ambiguous as Pink and White could be two possible answers depending on when they are born |
| Subjectivity | Contains clues that are highly subjective | "What's the name of Christopher Columbus's most famous ship?" Possible answers could be either Santa Maria, La Nina, Santa Clara. Also, as "Most famous" can mean many different things, the revised question could be "Which of Columbus's ships was stripped of its timbers to build a fort called La Navidad in northern Haiti?" |
| Ambiguity & Multiple acceptable answers | Can be answered with multiple answers | Nikolas Alexandrovitch Romanov, Nikolas II, Nikolai II Alexandrovich Romanov: all of these are acceptable as answers. |

Table 7: We list the problematic question types that we ask to annotate. The four types are illustrated with descriptions and examples to help them better understand each question, and help determine whether each question has good quality.

Figure 7: Writer Leaderboard in Interface

| Question Type | Adversarial Type |
| --- | --- |
| Composing seen clues | Contains clues that need to be integrated for the question to be answered |
| Logic and Calculation | Requires mathematical or logical operators |
| Multi-Step Reasoning | Requires multiple reasoning steps between entities. For eg: "A building dedicated to this man was the site of the "I Have A Dream" speech." A reasoning step is required to infer : "I have a dream" speech -> Lincoln Memorial -> Abraham Lincoln |
| Negation | Contains "not" or "non-" and "no" or any negation entities that may confuse the model to answer |
| Temporal Misalignment | Contains a specific year, month, or timely event that the model got confused about or does not know. |
| Location Misalignment | Contains a location that the model got confused about or does now know. |
| Commonsense Knowledge | Requires information that cannot be answered without commonsense |
| Domain Expert Knowledge | Requires information that cannot be answered without domain expert knowledge |
| Novel Clues | Contains information that exists in the question but is not required to answer. These confuse the models. |
| Crosslingual | Contains multilingual aspects that confuse the model. |

Table 8: We list adversarial types (techniques) to determine how each question is using them to stump the models. The annotators are given the description and examples to better understand the reasons why the models may have been stumped. They are expected to tag the examples with the model prediction and question.

| Question | Answer | Adversarial Type | Question Type | Grounding |
|---|---|---|---|---|
| What is a fourth of the 5th Bell number, often seen as an unlucky number? | 13/Thirteen | Logic & Calculation | Subjectivity | "Unlucky" is a subjective term. |
| What is the famous meme to come from The Last Dance? | and I took that personally | Commonsense Knowledge, Composing Seen Clues | Multiple Acceptable Answers | The meme can be referred to *many* titles: "Jordan's Cigar", "Jordan's Meme", "Laughing Jordan", and "Crying Jordan" |
| What substance can cause burns in its gaseous form, lead to vomiting and sweating in high doses, and is the main component by weight in acid rain? | Water | Logic & Calculation, Composing Seen Clues | Specificity | *Many substances* could cause these effects in the novel portion. |
| Name the title character of the 2024 Best Picture nominee about a fictional conductor who Leonard Bernstein mentored. | Lydia Tar | Temporal Misalignment, Composing Seen Clues | Factuality | 2024 Best Picture Nominee *cannot be factually identified* yet |
| The easternmost state in the U.S. has more than triple its population in lakes and it is known to have good salmon, which state is it? | Alaska | Multihop Reasoning&Location Misalignment | Subjectivity, Specificity | *Good salmon* is subjective, and *easternmost is misleading and it requires relative position* of the author, hence non-specific. |

Table 9: We annotated whether each question falls into which adversarial and question type. While being adversarial; some questions lack specificity and factuality. Other questions contained subjectivity and specificity.

| Question | Answer | Difficulty Rank |
|---|---|---|
| What is the name of the language which only has 45,900 speakers, allows for word-initial double consonants, and is the official language of an island country with the world's second-largest regional shark sanctuary? | Chuukese | 1 |
| What substance can cause burns in its gaseous form, lead to vomiting and sweating in high doses, and is the main component by weight in acid rain? | Water | 2 |
| A large portion of the sequence for the reference genome for the International Human Genome Sequencing Consortium in the human genome project came from a man from which US city? | Buffalo, New York | 3 |
| ⋮ | ⋮ | ⋮ |
| What year is the closest palindromic year to 2001? | 2022 | 88 |
| Which political party governs the country directly south of Botswana? | African National Congress | 89 |

Table 10: Question examples sorted by difficulty value from the IRT model

| Question | Answer | TFIDF | DPR | CHATGPT |
|---|---|---|---|---|
| Which political party governs the country directly south of Botswana? | African National Congress | A liberation movement known for its opposition to apartheid, it has governed the country since 1994, when the first post-apartheid election installed Nelson Mandela as President of South Africa. | *Rank 3* Partly due to its Tripartite Alliance with the SACP and the Congress of South African Trade Unions, it has retained a comfortable electoral majority at the national level and in most provinces, and has provided each of South Africa's five presidents since 1994. | The African National Congress (ANC) governs South Africa, which is directly south of Botswana. The ANC has been the ruling party in South Africa since the end of apartheid in 1994. |
| What company launched their first advertisement campaign during the 1984 Super Bowl that was inspired by George Orwell's dystopian novel "1984"? | Apple | From the introduction of the Macintosh in 1984, with the 1984 Super Bowl advertisement to the more modern Get a Mac adverts, Apple has been recognized for its efforts towards effective advertising and marketing for its products. | ***Incorrect Answer: Macintosh*** The first Mac was released in 1984, and was advertised with the highly-acclaimed "1984" ad. | ***Correct Answer: Apple*** The launch of Apple's '1984' ad during the Super Bowl is a well-known event in the advertising industry and is often cited as one of the most iconic ads of all time. |
| A play by Arthur Miller accounts what series of legal proceedings that accused over 200 individuals of consorting with the Devil in late 17th century New England? | Salem Witch Trials | Speakers at the ceremony in August included playwright Arthur Miller and Nobel Laureate Elie Wiesel. | *Rank 2* One other man, Giles Corey, was pressed to death after refusing to enter a plea, and at least five people died in jail. Arrests were made in numerous towns beyond Salem and Salem Village (known today as Danvers), notably Andover and Topsfield. | The question mentions a play by Arthur Miller, which is most likely 'The Crucible' - a dramatized retelling of the Salem Witch Trials. The question also mentions legal proceedings and accusations of consorting with the Devil, which are key elements of the Salem Witch Trials. |
| What is the title of the 2003 best-selling book that delves into the lives of two young boys in Afghanistan, exploring themes of friendship, betrayal, and redemption against the backdrop of political turmoil? | The Kite Runner | Because its themes of friendship, betrayal, guilt, redemption and the uneasy love between fathers and sons are universal, and not specifically Afghan, the book has been able to reach across cultural, racial, religious and gender gaps to resonate with readers of varying backgrounds. | *Rank 1* Hassan is a successful "kite runner" for Amir; he knows where the kite will land without watching it. | I predicted this answer because I recognized the themes of friendship, betrayal, and redemption as being central to The Kite Runner, which is a well-known and highly acclaimed novel set in Afghanistan. Additionally, the book was published in 2003 and was a best-seller, which further supports this answer. |
| This disney star had a show named after the state whose capital is Helena. This singer's most recent hit was a song about the brother of Chris Hemsworth. Liam Hemsworth, who is famously known for playing Josh Taylor, married which Disney star? | Miley Cyrus | While filming The Last Song, Cyrus began an on-again, off-again relationship with her co-star Liam Hemsworth in 2009. | *Rank 1* Cyrus, a daughter of country music singer Billy Ray Cyrus, emerged as a teen idol while portraying the title character of the Disney Channel television series Hannah Montana (2006–2011). | The Disney star with a show named after the state whose capital is Helena is Miley Cyrus, who starred in 'Hannah Montana'. Her most recent hit is 'Midnight Sky', which references Liam Hemsworth in the lyrics. Liam Hemsworth was previously married to Miley Cyrus. I predicted this answer because Miley Cyrus is a well-known Disney star who had a show named after a state capital, and she has been in the news recently for her music and personal life, including her relationship with Liam Hemsworth. |

Table 11: Example of Comparison between explanations that TFIDF, DPR, and CHATGPT provide for their respective answers. Some evidence shows similar content when predicting the correct answer. Some of the examples have one common clue that hints the models to their predictions, while for other examples, each model has their own groundings for their answers.